\documentclass[11pt,a4paper]{article}
\usepackage[hyperref]{emnlp-ijcnlp-2019}
\usepackage{latexsym}
\usepackage{times}
\usepackage{soul}
\usepackage{url}
\usepackage[utf8]{inputenc}
\usepackage{caption}
\usepackage{graphicx}
\usepackage{subfigure}
\usepackage{amsmath}
\usepackage{booktabs}
\usepackage{natbib}
\usepackage{xcolor}
\usepackage{fancyhdr,graphicx,amssymb}
\usepackage[ruled,vlined]{algorithm2e}
\usepackage{multirow}
\usepackage{url}
\usepackage{color}

\usepackage{tikz}
\usepackage{subfigure}
\usepackage{xcolor}
\usepackage{tcolorbox}

\usepackage{helvet}  
\usepackage{courier}  

\aclfinalcopy

\title{Improving Machine Reading Comprehension with Single-choice Decision and Transfer Learning}

\author{
  Yufan Jiang$^{1}$\thanks{Equal contribution. Correspondence to
  \{{\em frostwu}, {\em garyyfjiang}, {\em jennygong}, {\em huecheng}\}@{\em tencent.com}.},
  Shuangzhi Wu$^{1*}$,
  Jing Gong$^{2*}$,
  Yahui Cheng$^{2*}$, \\
  \textbf{Peng Meng$^{2}$},
  \textbf{Weiliang Lin$^{2}$},
  \textbf{Zhibo Chen$^{2}$}
  \textbf{and Mu Li$^{1}$}\\
  $^{1}$Tencent Cloud Xiaowei\\ 
  $^{2}$Tencent Cloud TI-ONE \\
  {\tt
        \{frostwu,garyyfjiang,jennygong,huecheng\}@tencent.com 
  }\\
  {\tt
        \{pengmeng,weilianglin,ruibobchen,ethanlli\}@tencent.com 
  }\\
}

\date{}

\begin{document}
\maketitle

\begin{abstract}
Multi-choice Machine Reading Comprehension (MMRC) aims to select the correct answer from a set of options based on a given
passage and question. Due to task specific
of MMRC, it is non-trivial to transfer knowledge from other MRC tasks such as SQuAD,
Dream. In this paper, we simply reconstruct
multi-choice to single-choice by training a binary classification to distinguish whether a
certain answer is correct. Then select the option with the highest confidence score. We
construct our model upon ALBERT-xxlarge
model and estimate it on the RACE dataset.
During training, We adopt AutoML strategy
to tune better parameters. Experimental results show that the single-choice is better than
multi-choice. In addition, by transferring
knowledge from other kinds of MRC tasks,
our model achieves a new state-of-the-art results in both single and ensemble settings.

\end{abstract}

\section{Introduction}

The last several years have seen a land rush in research on machine reading (MRC) comprehension and various dataset have been proposed such as SQuAD1.1, SQuAD2.0, NewsQA and CoQA \cite{rajpurkar2016squad,trischler2016newsqa,reddy2019coqa}. Different from the above which are extractive MRC, RACE is a multi-choice MRC dataset (MMRC) proposed by \cite{lai2017race}. RACE was extracted from middle and high school English examinations in China. Figure 1 shows an example passage and two related questions from RACE. The key difference between RACE and previously released machine comprehension datasets is that the answers in RACE often cannot be directly extracted from the passages, as illustrated by the two example questions (Q1 \& Q2) in Table \ref{tab:example}. Thus, answering these questions needs inferences.

\begin{table}[t!]
  \small
  \begin{center}
  \begin{tabular}{p{7.2cm}}
  \toprule
  \textbf{Passage}: For the past two years, 8-year-old Harli Jordean from Stoke Newington, London, has been selling marbles . His successful marble company, Marble King, sells all things marble-related - from affordable tubs of the glass playthings to significantly expensive items like Duke of York solitaire tables - sourced, purchased and processed by the mini-CEO himself. "I like having my own company. I like being the boss," Harli told the Mirror....Tina told The Daily Mail. "At the moment he is annoying me by creating his own Marble King marbles - so that could well be the next step for him." \\
  \midrule
  \textbf{Q1}: Harli's Marble Company became popular as soon as he launched it because \_\_\_ .     \\
  \textbf{A}: it was run by "the world's youngest CEO"      \\
  \textbf{B: it filled the gap of online marble trade }     \\
  \textbf{C}: Harli was fascinated with marble collection   \\
  \textbf{D}: Harli met the growing demand of the customers \\
  \midrule
  \textbf{Q2}: How many mass media are mentioned in the passage?    \\
  \textbf{A}: One      \\
  \textbf{B}: Two     \\
  \textbf{C: Three}   \\
  \textbf{D}: Four \\
  \bottomrule
  \end{tabular}
  \end{center}
  \caption{An example passage and two related multi-choice questions. The ground-truth answers are in \textbf{bold}.}
  \label{tab:example}
\end{table}


Recently, pretrained language models (LM) such as BERT \cite{devlin2018bert}, RoBERTa \cite{liu2019roberta}, ALBERT \cite{lan2019albert} have achieved great success on MMRC tasks. Notably, Megatron-LM \cite{shoeybi2019megatron} which is a 48 layer BERT with 3.9 billion parameters yields the highest score on the RACE leaderboard in both single and ensemble settings. The key point to model MMRC is: first encode the context, question, options with BERT like LM, then add a matching network on top of BERT to score the options. Generally, the matching network can be various \cite{ran2019option, zhang2020dcmn, zhu2020duma}. \newcite{ran2019option} proposes an option comparison network (OCN) to compare options at word-level to better identify their correlations to help reasoning. \newcite{zhang2020dcmn} proposes a dual co-matching network (DCMN) which models the relationship among passage, question and answer options bidirectionally. All these matching networks show promising improvements compared with pretrained language models. One point they have in common is that the answer together with the distractors are jointly considered which we name multi-choice models. We argue that the options can be concerned separately for two reasons, 1) when human works on MMRC problem, they always consider the options one by one and select the one with the highest confidence. 2) MMRC suffers from the data scarcity problem. Multi-choice models are inconvenient to take advantage of other MRC dataset. 

In this paper, we propose a single-choice model for MMRC. Our model considers the options separately. The key component of our method is a binary classification network on top of pretrained language models. For each option of a given context and question, we calculate a confidence score. Then we select the one with the highest score as the final answer. In both training and decoding, the right answer and the distractors are modeled independently. Our proposed method gets rid of the multi-choice framework, and can leverage amount of other resources. Taking SQuAD as an example, we can take a context, one of its question and the corresponding answer as a positive instance for our classification with golden label 1. In this way many QA dataset can be used to enhance RACE. Experimental results show that single-choice model performs better than multi-choice models, in addition by transferring knowledge from other QA dataset, our single model achieves 90.7\% and ensemble model achieves 91.4\%, both are the best score on the leaderboard. 

\begin{figure}[t!]
  \centering
  \tikzstyle{Pnode} = [rounded corners=1pt,inner sep=4pt,minimum height=1.3em,minimum width=3em,draw,thick,fill=red!20]
  \tikzstyle{Qnode} = [rounded corners=1pt,inner sep=4pt,minimum height=1.3em,minimum width=1.6em,draw,thick,fill=blue!20]
  \tikzstyle{Anode} = [rounded corners=1pt,inner sep=4pt,minimum height=1.3em,minimum width=1.6em,draw,thick,fill=lightgray!30]
  \tikzstyle{encnode} = [rounded corners=1pt,inner sep=4pt,minimum height=1.5em,minimum width=1.3em,draw,thick,fill=lightgray!40]
  \tikzstyle{labelnode} = [rounded corners=1pt,inner sep=4pt,minimum height=0.8em,minimum width=0.8em,draw,thick,fill=lightgray!20]
  \tikzstyle{standard} = [rounded corners=2pt,thick]
  \centering
    \begin{tikzpicture}
    \node [Pnode,anchor=west,draw=black!40] (p4) at (0,0) {\tiny{$P$}};
    \node [Pnode,anchor=south west,draw=black!60,inner sep=4pt] (p3) at ([shift={(-0.5em,-0.5em)}]p4.south west) {\tiny{$P$}};
    \node [Pnode,anchor=south west,draw=black!80,inner sep=4pt] (p2) at ([shift={(-0.5em,-0.5em)}]p3.south west) {\tiny{$P$}};
    \node [Pnode,anchor=south west,draw=black!90,inner sep=4pt] (p1) at ([shift={(-0.5em,-0.5em)}]p2.south west) {\tiny{$P$}};
    
    \node [Qnode,anchor=west,draw=black!40] (q4) at ([xshift=-0.05em]p4.east) {\tiny{$Q$}};
    \node [Qnode,anchor=west,draw=black!60] (q3) at ([xshift=-0.05em]p3.east) {\tiny{$Q$}};
    \node [Qnode,anchor=west,draw=black!80] (q2) at ([xshift=-0.05em]p2.east) {\tiny{$Q$}};
    \node [Qnode,anchor=west,draw=black!90] (q1) at ([xshift=-0.05em]p1.east) {\tiny{$Q$}};
    
    \node [Anode,anchor=west,draw=black!40,fill=lightgray!30] (a4) at ([xshift=-0.05em]q4.east) {\tiny{$A_4$}};
    \node [Anode,anchor=west,draw=black!60,fill=yellow!30] (a3) at ([xshift=-0.05em]q3.east) {\tiny{$A_3$}};
    \node [Anode,anchor=west,draw=black!80,fill=green!30] (a2) at ([xshift=-0.05em]q2.east) {\tiny{$A_2$}};
    \node [Anode,anchor=west,draw=black!90,fill=orange!30] (a1) at ([xshift=-0.05em]q1.east) {\tiny{$A_1$}};
    
    \node [encnode,anchor=west] (enc1) at ([xshift=2em]a2.east) {\footnotesize{Enc}};
    \node [encnode,anchor=west] (dec1) at ([xshift=1.5em]enc1.east) {\footnotesize{Dec}};
    
    \node [labelnode,anchor=west] (label1) at ([xshift=-2em, yshift=3em]dec1.west) {\tiny{0}};
    \node [labelnode,anchor=west] (label2) at ([xshift=-0.1em]label1.east) {\tiny{0}};
    \node [labelnode,anchor=west] (label3) at ([xshift=-0.1em]label2.east) {\tiny{1}};
    \node [labelnode,anchor=west] (label4) at ([xshift=-0.1em]label3.east) {\tiny{0}};
    
    \node [anchor=west] (loss1) at ([xshift=1.5em]dec1.east) {\footnotesize{Loss}};
    
    \draw [->,thick] ([xshift=0.3em, yshift=-0.5em]a1.east) -- ([xshift=-0.1em,yshift=-0.1em]enc1.west);
    \draw [->,thick] ([xshift=0.1em]a4.east) -- ([xshift=-0.1em,yshift=0.1em]enc1.west);
    \draw [->,thick] ([xshift=0.1em]enc1.east) -- ([xshift=-0.1em]dec1.west);
    \draw [->,thick] ([xshift=0.1em]dec1.east) -- ([xshift=-0.1em]loss1.west);
    
    \draw [->,standard, thick] ([xshift=0.1em]label4.east) -- ([xshift=1.5em]label4.east) -- ([xshift=1.5em, yshift=-2.7em]label4.east);
     
    \draw [->,thick,standard,red] ([xshift=0.1em]loss1.east) -- ([xshift=0.6em]loss1.east) -- ([xshift=0.6em,yshift=-2.0em]loss1.east) -- ([xshift=-3em,yshift=-2.0em]loss1.east);
    
    \node [anchor=north] (a) at ([yshift=-2.3em]enc1.south) {\small (a) Multi-choice Model};
    \node [anchor=south] (bp) at ([xshift=2em,yshift=3em]p4.north) {\small{: Back Propagation} };
    
    \draw [->,thick,standard,red] ([xshift=-1.5em]bp.west) -- ([xshift=0.1em]bp.west);
    
    \node [Pnode,anchor=north,draw=black!40] (p5) at ([yshift=-9em]p4.south) {\tiny{$P$}};
    \node [Pnode,anchor=south west,draw=black!60,inner sep=4pt,fill=orange!50] (p6) at ([shift={(-0.5em,-0.5em)}]p5.south west) {\tiny{$P$}};
    \node [Pnode,anchor=south west,draw=black!80,inner sep=4pt,fill=yellow!40] (p7) at ([shift={(-0.5em,-0.5em)}]p6.south west) {\tiny{$P$}};
    \node [Pnode,anchor=south west,draw=black!90,inner sep=4pt,fill=blue!20] (p8) at ([shift={(-0.5em,-0.5em)}]p7.south west) {\tiny{$P_3$}};
    
    \node [Qnode,anchor=west,draw=black!40,fill=green!30] (q5) at ([xshift=-0.05em]p5.east) {\tiny{$Q$}};
    \node [Qnode,anchor=west,draw=black!60] (q6) at ([xshift=-0.05em]p6.east) {\tiny{$Q$}};
    \node [Qnode,anchor=west,draw=black!80,fill=lightgray!20] (q7) at ([xshift=-0.05em]p7.east) {\tiny{$Q$}};
    \node [Qnode,anchor=west,draw=black!90,fill=red!20] (q8) at ([xshift=-0.05em]p8.east) {\tiny{$Q_2$}};
    
    \node [Anode,anchor=west,draw=black!40,fill=lightgray!30] (a5) at ([xshift=-0.05em]q5.east) {\tiny{$A_4$}};
    \node [Anode,anchor=west,draw=black!60,fill=yellow!30] (a6) at ([xshift=-0.05em]q6.east) {\tiny{$A_3$}};
    \node [Anode,anchor=west,draw=black!80,fill=green!30] (a7) at ([xshift=-0.05em]q7.east) {\tiny{$A_2$}};
    \node [Anode,anchor=west,draw=black!90,fill=orange!30] (a8) at ([xshift=-0.05em]q8.east) {\tiny{$A_1$}};
    
    \node [encnode,anchor=west] (enc2) at ([xshift=2em]a7.east) {\footnotesize{Enc}};
    \node [encnode,anchor=west] (dec2) at ([xshift=1.5em]enc2.east) {\footnotesize{Dec}};
    
    \node [anchor=west] (loss2) at ([xshift=1.5em]dec2.east) {\footnotesize{Loss}};
    
    \node [labelnode,anchor=west] (label5) at ([xshift=0em, yshift=4em]dec2.west) {\tiny{0}};
    \node [labelnode,anchor=west] (label51) at ([xshift=-0.1em]label5.east) {\tiny{1}};
    \node [labelnode,anchor=south west] (label6) at ([shift={(-0.5em,-0.5em)}]label5.south west) {\tiny{0}};
    \node [labelnode,anchor=west] (label61) at ([xshift=-0.1em]label6.east) {\tiny{1}};
    \node [labelnode,anchor=south west] (label7) at ([shift={(-0.5em,-0.5em)}]label6.south west) {\tiny{1}};
    \node [labelnode,anchor=west] (label71) at ([xshift=-0.1em]label7.east) {\tiny{0}};
    \node [labelnode,anchor=south west] (label8) at ([shift={(-0.5em,-0.5em)}]label7.south west) {\tiny{0}};
    \node [labelnode,anchor=west] (label81) at ([xshift=-0.1em]label8.east) {\tiny{1}};
    
     \draw [->,thick] ([xshift=0.3em, yshift=-0.5em]a8.east) -- ([xshift=-0.1em,yshift=-0.1em]enc2.west);
    \draw [->,thick] ([xshift=0.1em]a5.east) -- ([xshift=-0.1em,yshift=0.1em]enc2.west);
    \draw [->,thick] ([xshift=0.1em]enc2.east) -- ([xshift=-0.1em]dec2.west);
    \draw [->,thick] ([xshift=0.1em]dec2.east) -- ([xshift=-0.1em]loss2.west);
    
    \node [anchor=north] (b) at ([yshift=-2.3em]enc2.south) {\small (b) Single-choice Model};
    
    \draw [->,thick,standard,red] ([xshift=0.1em]loss2.east) -- ([xshift=0.6em]loss2.east) -- ([xshift=0.6em,yshift=-2.0em]loss2.east) -- ([xshift=-3em,yshift=-2.0em]loss2.east);
    
    \draw [->,standard, thick] ([xshift=1em]label71.east) -- ([xshift=2.5em]label71.east) -- ([xshift=2.5em, yshift=-2.7em]label71.east);

    \end{tikzpicture}
    \caption{An overview of Standard Model and Single-choice Model.}
\end{figure}

\section{Task Description}

Multi-choice MRC (MMRC) can be represented as a triple $<P, Q, A>$, where $P=s_1,s_2,...,s_m$ is an article consist of multiple sentences $s$, $Q$ is a question asked upon the article and $A=\{A_1,...A_n\}$ is a set of candidate answers. Only one answer in $A$ is correct and others are distractors. The purpose of the MMRC is to select the right one. RACE is one kind of MMRC task, which is created by domain experts to test students' reading comprehension skills, consequently requiring non-trivial reasoning techniques. Each article in RACE has several questions and the questions always have 4 candidate answers, one answer and three distractors.

\section{Methods}

Previous works have verified the effectiveness of Pretrained language models such as BERT, XLNet,Roberta and Albert in Multi-choice MRC tasks. Pretrained language models are used as encoder to get the global context representation. After that, a decoder is employed to find the correct answer given all the information contained in the global representation.
Let $P$, $Q$, and $\{A_1,...A_n\}$ denote the passage, question and option set separately.
The input of Pretrained encoder is defined as $(P \oplus Q \oplus A_i)$, the concatenation of $P$, $Q$ and $A_i$ one of the option in candidate set.
Moreover, for the same question, the inputs with different options are concatenated  together as a complete training sample, which is more intuitive and similar to human that select the correct answer compared with other options. After encoding all the inputs for a single question, the contextual representations $T = \{T_{CLS1},...,T_{CLSn}\}$ is used to classify which is the correct answer given passage and question(see Figure2(a)). An single full connection layer is added to compute the probability $p(\{A_1,...A_n\}|P,Q)$ for all the answers and the ground truth $y$ is the index of correct answer in candidates. $T\in \mathbb R^{n\times h}$, $n$ denotes the number of the options in candidate set. We define the score to be:
\begin{equation}
p(\{A_1,...A_n\}|P,Q) = \sigma (WT + b)
\end{equation}
where $W\in\mathbb R^{h\times1}$ is the weight and $b$ is the bias. Parameter matrices are finetuned based on pretrained language model with the cross entropy loss function which is formulated as:
\begin{eqnarray}
\text{loss} &=& -\sum y \log (p)
\end{eqnarray}

\subsection{Single-choice Model}
As all input sequences with the same passage and question are tied together, each training sample contain much duplicate content. For example, the passage with multiple sentences repeat $n$ times in a single training sample which may degrade the diversity in each training step.
Moreover, this method need to fix the data format that each question must have the same number of options which is also inconvenient to take advantage of other MRC datasets.

Alternatively, we reconstruct the multi-choice to single-choice.
We just need to distinguish whether the answer is correct without considering other options in the candidate set. By this way, we keep the diversity in training batches and relax the constraints on multi-choice framework.

Instead of concatenate all inputs with the same question together, we just  encode a single input and use its contextual representations $T_{CLSi}$ to classify whether the answer is correct (see Figure2(b)). The ground truth is $y \in \{0,1 \}$, Thus we re-define the score  $g(P,Q,A_{i})$ as:
\begin{equation}
g(P,Q,A_{i}) = \sigma (WT_{CLSi} + b)
\end{equation}
where $W\in\mathbb R^{h\times label}$. Correspondingly, the cross entropy loss function can be re-formulated as:
\begin{eqnarray}
\text{loss} &=& -\sum y \log (g(P,Q,A_{i})) \nonumber \\
&+& (1-y) \log (1-g(P,Q,A_{i})) 
\end{eqnarray}

In the end, to get the correct answers, we select the top-n answers with respect to score. Here $n$ denotes the number of correct answers. E.g., $n$=1 in RACE. 

\begin{figure}[htb!]
\includegraphics[width=0.5\textwidth]{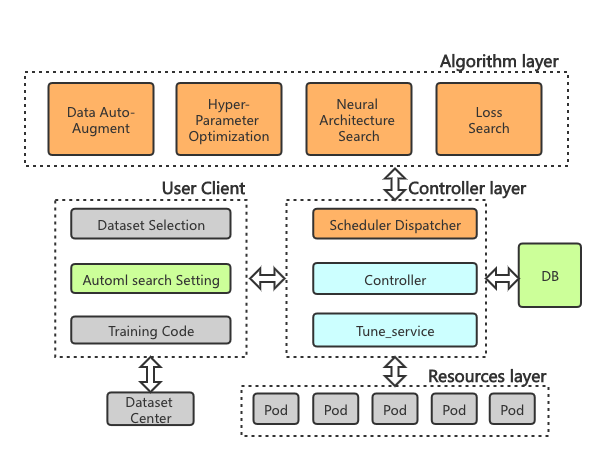}
\flushleft
\caption{Overview of our AutoML architecture.}
\label{fig:automl}
\end{figure}

\subsection{Transfer Learning}

In this section, We propose a simple yet effective strategy to transfer knowledge from other QA dataset. As the single-choice model relax the constraints on multi-choice framework, more QA datasets such as SQuAD2.0, ARC, CoQA and DREAM can be used to enhance RACE. It consists of three steps:

(1) we preprocess data with different formats to the same input type as mentioned in section 3. For multiple-choice MRC datasets, like DREAM and ARC, we concatenate each option with corresponding context and question.
And for extractive MRC datasets like SQuAD2.0 and CoQA, we take the context (passage or dialog), one of its question and corresponding answer as a positive instance for the binary classification.

(2) We collect and corrupt the preprocessed data from different QA datasets
and then train the binary classification on this mixed data. We find that the model benefits a lot from the large amount of MRC datasets.

(3) Finally, we further finetune the model from step 2 on the raw RACE data to adapt the model parameters to the task.
 
\begin{table*}[t]
    \centering
    \begin{tabular}{lcccccc}
    \hline 
        Models & RACE & DREAM & SQuAD2.0 & CoQA & ARC & Crawl \\ \hline
        \#Article  & 27,933 & 6,444 & 130,319 & 8,399 & - &  - \\
        \#Question  & 97,687 & 10,197 & - & 127,000 & 7787 &  -  \\
        \#Answer per Question  & 4 & 3 & 1 & - & 4 &  -  \\
        \#Word per Article & 321.9 & 85.9 & - & 271 & - &  -  \\
        \#Instance  & 351,464 & 15,470 & 86,835 & 90,000 & 20,784 &  446,095  \\
        \hline
    \end{tabular}
    \caption{Details of different MRC resources. ``\#Instance" refers to the number of true training samples built by different resources.}
    \label{tab:data}
\end{table*}
 
\section{Tencent TI-ONE Platform\footnote{https://cloud.tencent.com/product/tione}}

Our systems are built on TI-ONE, which is a deep learning platform built on the Tencent GPU Cloud. It is an industrial platform with advanced technologies and rich features including  popular deep learning frameworks, automated machine learning, large scale distributed training as well as service platforms. 
We adopt the AutoML algorithm to select better hyper parameters and accelerate the process by distributed training.

\subsection{AutoML}

Finetuning pretrained language models on downstream tasks is sensitive to the selection of hyper parameters. A good set of hyper parameter affects the final performance to a great extent. However, it is impossible to manually search an optimize set from the huge amount of hyper parameter combinations. To alleviate this problem, we take advantage of automated machine learning (AutoML)\cite{he2019automl,zoller2019survey,elshawi2019automated} to automatically adapt hyper parameters. 

Our AutoML system named TianFeng is a lightweight, extensible, and easy-to-use framework. TianFeng incorporates most current state-of-the-art algorithms and can make good use of resources. It consists of three parts, internal layers, algorithm layer, controller layer and resources layer, as shown in Figure\ref{fig:automl}.

(1) Algorithm layer. The algorithm layer has 4 sub modules, which are used to search model parameters from different perspectives.


(2) Controller layer. Responsible for docking Client, issuing algorithm logic and exception handling.

(3) Resources layer. Effective management of resource pools.

In general, the controller layer receives request from client and select a proper algorithm from algorithm layer. Then uses the idle GPU computing resources in the resource pool to perform multiple tasks in parallel.

\subsection{Distributed Training}

Due to the huge amount parameters of pretrained language models, it is very time-consuming to conduct AutoML training. We take advantage of the distributed training techniques on Tencent Cloud TI-ONE platform to make the training more efficient. The mixed precision training is used, which greatly accelerate training on single-machine. When training on multiple machines, TI-ONE's fast communication framework can fully leverage more GPUs. 
Two main advantages on multi-machine communication are: 1) Use optimized all-reduce algorithm and multi-stream to make full use of the bandwidth of VPC network 2) Support gradient fusion of multiple strategies to improve communication efficiency.
Our best model is trained on 4 machines with 32 V100 GPUs. The training time can be shortened to 33 percent of the original single machine with 8 cards.


\section{Experiments}

\subsection{Dataset}

\paragraph{RACE} RACE \cite{lai2017race} is dataset collected from middle and high school Englis exams in China. RACE has a wide variety of question type such as summarization, inference, deduction and context matching. It contains articles from multiple domains (i.e. news, ads, story) and most of the questions need reasoning. 

In the transfer learning stage, we also consider other MRC tasks. Specifically, we consider SQuAD2.0 \cite{rajpurkar2016squad}, ARC \cite{Clark2018ThinkYH}, CoQA \cite{reddy2019coqa} and DREAM \cite{sundream2018}. We give a brief description of these datasets.

\paragraph{SQuAD2.0 and CoQA} SQuAD2.0 and CoQA are extractive MRC tasks, the articles of which are wiki passages and dialogs. Their questions do not have candidate answers, instead participants are asked to extract the answer from the passage.

\paragraph{ARC} ARC is the largest public-domain multiple-choice dataset that consist of natural and grade-school science question. It is partitioned into a Challenge Set and an Easy set.

\paragraph{DREAM} DREAM is multiple-choice dialogue-based Reading comprehension examination dataset. It article is a dialog and each question has only three options.

Although we have transferred as much data as we can, the MMRC task still suffers data insufficiency problem. Thus we crawl different kind of MRC data from website. Table \ref{tab:data} lists all the resources we use.

\begin{table}[t]
    \small
    \centering
    \begin{tabular}{ll}
    \hline 
        Models & Test \\ \hline
        Roberta \cite{liu2019roberta} &    83.2   \\
        ALBERT (single) \cite{lan2019albert} &    86.5   \\
        ALBERT (ensemble) \cite{lan2019albert} &   89.4   \\
        ALBERT + DUMA (single) \cite{zhu2020duma} &   88.0   \\
        ALBERT + DUMA (ensemble) \cite{zhu2020duma} &  89.8   \\
        Megatron-BERT (single) \cite{shoeybi2019megatron}   &  \bf 89.5   \\
        Megatron-BERT (ensemble) \cite{shoeybi2019megatron}  &  \bf 90.9   \\
        \hline
        ALBERT baseline        & 87.1   \\
        ALBERT single-choice   & 87.9   \\
        + transfer learning   &  88.3   \\
        + AutoML  & \bf 90.0   \\
        + Crawled corpus  & \bf 90.7   \\
        Ensemble & \bf 91.4   \\
        \hline
    \end{tabular}
    \caption{Results on RACE dataset.}
    \label{tab:main-result}
\end{table}

\subsection{Experimental Settings}

Our implementation was based on Transformers\footnote{https://github.com/huggingface/transformers}. We use the ALBERT-xxlarge as encoder. For hyper parameters, we follow Table 15 in \cite{lan2019albert}, except that we set the learning rate to 1e-5 and the warmup steps to 2000. Because we find this is better for the huggingface ALBERT-xxlarge model. After adding the other resources, we do not use a fixed ``Training Steps", the training steps after two epochs and the warm up step is 10\% of the total training steps. All the models are trained on 8 nVidia V100 GPUs. The training takes about 2 days. 

\paragraph{Baseline} Our baseline is the original huggingface ALBERT-xxlarge model with the default multi-choice strategy. The hyper parameters follow the description above. In addition, we compare our model with many other public results from both papers or the leaderboard.

\subsection{Results}

Table \ref{tab:main-result} shows the results of our models and the baselines. The top part of the table lists the results from the current leaderboard \footnote{http://www.qizhexie.com/data/RACE\_leaderboard.html} and papers. Megatron-BERT \cite{shoeybi2019megatron} achieves the best single and ensemble results. It is a variant of BERT\cite{devlin2018bert} with 3.9 billion parameters which is almost 40 times bigger than ALBERT-xxlarge. 

The results of our models are listed below the table. Our ALBERT baseline yields better result than original ALBERT due to the different choice of hyper parameters illustrated in 5.2 showing that the task is sensitive to hyper parameters. Compared with the baseline, our single-choice model achieves 0.8 more score, which shows that single-choice is better than multi-choice under the ALBERT-xxlarge model. After transferring knowledge from other MRC dataset, we get another 0.4 more score. With the help of autoML, our single model achieves 90\% which surpasses Megatron-BERT \cite{shoeybi2019megatron} and becomes the new state-of-the-art single model results. When adding the web crawl corpus into transfer learning, our single model get the final score as high as 90.7\%. This illustrates that single-choice model is easy to incorporate other resources and we achieve this by a simple transfer learning strategy. Our ensemble model gets the best score of 91.4\%. 

\section{Conclusion}
In this paper, we propose a single-choice model for MMRC that consider the options separately. Experiments results demonstrate that our method achieves significantly improvements and by taking advantage of other MRC datasets, we achieve a new state-of-the-art performance. We plan to consider the difference between two methods and if we can combine them together in future study.

\bibliography{emnlp-ijcnlp-2019}
\bibliographystyle{acl_natbib}

\end{document}